\documentclass[letterpaper]{article} %
\usepackage{aaai2026}  %
\usepackage{times}  %
\usepackage{helvet}  %
\usepackage{courier}  %
\usepackage[hyphens]{url}  %
\usepackage{graphicx} %
\usepackage{natbib}  %
\usepackage{caption} %
\usepackage{algorithm}
\usepackage{algorithmic}

\usepackage{newfloat}
\usepackage{listings}
\DeclareCaptionStyle{ruled}{labelfont=normalfont,labelsep=colon,strut=off} %
\floatstyle{ruled}
\newfloat{listing}{tb}{lst}{}
\floatname{listing}{Listing}
\usepackage{amssymb}
\usepackage{amsmath}
\usepackage{multirow}
\usepackage{booktabs}

\title{VGGTFace: Topologically Consistent Facial Geometry Reconstruction in the Wild}

\author {
    Xin Ming,
    Yuxuan Han\thanks{Corresponding author.},
    Tianyu Huang,
    Feng Xu
}
\affiliations {
    BNRist and School of Software, Tsinghua University \\
    1729406968@qq.com, hanyx22@mails.tsinghua.edu.cn, huang-ty21@mails.tsinghua.edu.cn,
    xufeng2003@gmail.com
}

\usepackage{bibentry}

\begin{document}

\maketitle

\begin{figure*}[t]
    \centering
    \includegraphics[width=\textwidth]{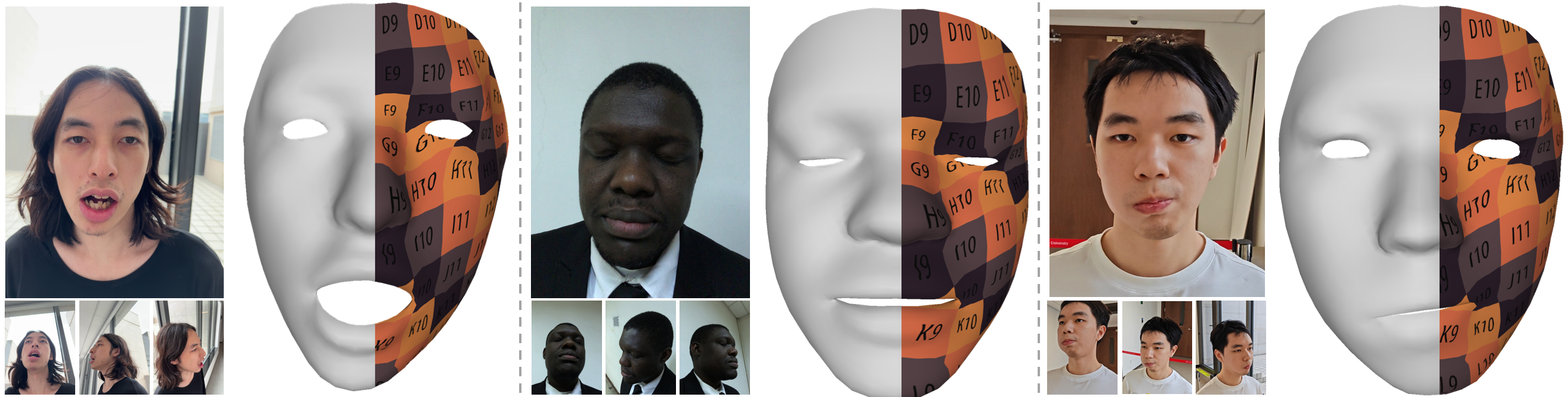}
    \caption{
        VGGTFace. Given in-the-wild multi-view images (4 of 16 images are shown here) captured by everyday users as input, our method can reconstruct a topologically consistent mesh from these inputs in 10 seconds. Our method demonstrates strong generalization ability across lighting conditions of the capture site, facial expressions, and ethnic groups, opening the door for everyday users to scan themselves into the digital world. By innovatively applying the point map as the facial geometry representation, our method can reconstruct person-specific facial traits, \emph{e.g.} asymmetric expressions, with high fidelity.
    }
    \label{Fig:teaser}
\end{figure*}

\begin{abstract}
Reconstructing topologically consistent facial geometry is crucial for the digital avatar creation pipelines.
Existing methods either require tedious manual efforts, lack generalization to in-the-wild data, or are constrained by the limited expressiveness of 3D Morphable Models.
To address these limitations, we propose VGGTFace, an automatic approach that innovatively applies the 3D foundation model, \emph{i.e.} VGGT, for topologically consistent facial geometry reconstruction from in-the-wild multi-view images captured by everyday users.
Our key insight is that, by leveraging VGGT, our method naturally inherits strong generalization ability and expressive power from its large-scale training and point map representation.
However, it is unclear how to reconstruct a topologically consistent mesh from VGGT, as the topology information is missing in its prediction.
To this end, we augment VGGT with Pixel3DMM for injecting topology information via pixel-aligned UV values. 
In this manner, we convert the pixel-aligned point map of VGGT to a point cloud with topology.
Tailored to this point cloud with known topology, we propose a novel Topology-Aware Bundle Adjustment strategy to fuse them, where we construct a Laplacian energy for the Bundle Adjustment objective.
Our method achieves high-quality reconstruction in 10 seconds for 16 views on a single NVIDIA RTX 4090. 
Experiments demonstrate state-of-the-art results on benchmarks and impressive generalization to in-the-wild data. 
Code is available at https://github.com/grignarder/vggtface.
\end{abstract}

\section{Introduction}
Reconstructing facial geometry with consistent topology is crucial for digital avatar pipelines~\cite{alexander2009digital,alexander2013digital}, as it enables dense mesh correspondence and transferable animation, rigging, and texture editing. Traditional workflows first reconstruct a detailed facial scan and then enforce topological consistency by fitting a template mesh~\cite{egger20203d,riviere2020single}. However, achieving high quality typically requires substantial manual effort—such as correspondence marking and parameter tuning—making these pipelines difficult to scale to everyday users.

There exist some pioneering works that attempt to reconstruct a topologically consistent face mesh automatically. 
Some works take multi-view images as input and use neural networks to directly predict the results~\cite{li2024grape,Bolkart2023Tempeh,li2021tofu}.
Training on studio-captured data, these methods can only infer images captured in a similar studio.
However, they cannot generalize to in-the-wild faces captured by everyday users, as the lighting conditions and camera configurations differ significantly from their training data.
Another group of works exploits the 3D Morphable Face Model (3DMM) for topologically consistent facial geometry reconstruction~\cite{tewari2019fml,deng2019accurate,DECA:Siggraph2021,wang20243d}.
Building on top of the 3DMM prior, these methods can leverage large-scale face image or video datasets for training. 
However, their expressiveness is confined to the limited 3DMM space.
Although some works propose more powerful representations, such as 3DMM with per-vertex displacement~\cite{lei2023hierarchical} or corrective basis~\cite{bai2020deep}, their dependency on 3DMM inevitably constrains the reconstruction quality, as shown by \citet{li2021tofu} and our experiments.

Recently, 3D foundation models, such as VGGT~\cite{wang2025vggt}, have reshaped the field of geometry reconstruction. 
Given multi-view images captured in the wild, VGGT can infer the pixel-aligned point map (\emph{i.e.} the 3D position of each pixel) and camera intrinsic and extrinsic for each view in a single forward pass.
By training on large-scale datasets, VGGT is robust to diverse lighting conditions and camera configurations that appear in the data captured by everyday users. 
Meanwhile, the point map representation adopted by VGGT has enough expressiveness to capture high-quality facial geometry.
These nice properties motivate us to apply VGGT for topologically consistent facial geometry reconstruction.
However, it is unclear how to fulfill this as topology information is missing in the VGGT's prediction.

To address this, we propose VGGTFace, a novel method for topologically consistent facial geometry reconstruction from in-the-wild multi-view images captured by everyday users.
To best maintain the generalization ability of VGGT, we freeze its pretrained weights to infer pixel-aligned point maps and camera parameters from the captured multi-view face images.
Next, we augment the point maps with pixel-aligned UV values using Pixel3DMM~\cite{giebenhain2025pixel3dmm}, where the UV values indicate the corresponding positions on a template face mesh.
As each vertex on the template mesh has a unique pair of UV values, we assign the pixel with the same UV values as the 2D projected point of the vertex on this view.
In this manner, we get the correspondence information to build the topologically consistent face mesh. 
However, as neither VGGT nor Pixel3DMM is perfectly accurate, the 3D position of the same template vertex computed from different views is inconsistent. 
Fusing them to minimize errors and thus reconstruct a topologically consistent mesh is still a challenging task.

To solve this problem, our key insight is to regard the corresponding pixel of the template vertex at each view as tracks (the target projected 2D points of a 3D point on different views~\cite{hartley2003multiple}) and formulate the VGGT point cloud fusion as a Bundle Adjustment problem~\cite{triggs1999bundle}.
Specifically, we simultaneously optimize camera parameters and the fused point cloud to minimize the re-projection error \emph{w.r.t} projected distance to the tracks.
However, these tracks also contain errors, thus leading to a noisy fused point cloud after Bundle Adjustment (BA).
To address this, existing works typically require tedious parameter tweaking to filter out the influence of low-quality tracks~\cite{schoenberger2016mvs,schoenberger2016sfm}.
In our task, as the VGGT point cloud has one-to-one correspondences to the template mesh after being augmented with UV coordinates, we can exploit the topology information of the template mesh to better perform BA.
To be specific, we construct a Laplacian energy~\cite{Nicolet2021Large} to the BA objective as a regularization to effectively overcome the low-quality tracks.
In addition, the Laplacian energy can provide gradients to every 3D point, effectively regularizing unconstrained points that have less than two valid tracks in the BA system.
Note that previous BA methods cannot use Laplacian energy as their point cloud does not have topology.
Thus, we term our technique Topology-Aware Bundle Adjustment (TopBA).
After the TopBA process, we directly connect the fused point cloud to obtain the result.
Thanks to the highly optimized BA solver~\cite{Agarwal_Ceres_Solver_2022}, our method can reconstruct a high-quality topologically consistent mesh from 16 views in 10 seconds on a single NVIDIA RTX 4090.
In conclusion, our main contributions include:
\begin{itemize}
    \item We propose a novel method that can reconstruct topologically consistent facial geometry from in-the-wild multi-view images captured by everyday users in 10 seconds.
    \item We propose a novel Topology-Aware Bundle Adjustment (TopBA) technique to robustly distill a topologically consistent mesh from the raw point cloud predicted by VGGT.
    \item We demonstrate state-of-the-art geometry reconstruction results on various benchmarks and impressive generalization ability on in-the-wild captured data.
\end{itemize}

We hope our method provides a starting point for applying 3D foundation models to face-related tasks, helping overcome the scarcity of 3D facial data and enabling high-quality reconstruction for everyday users in the wild.

\section{Related Works}

\subsection{Monocular Face Reconstruction}
As reconstructing geometry from a single image is an ill-posed problem, most methods rely on the 3DMM prior~\cite{DBLP:conf/siggraph/BlanzV99,zollhofer2018state}.
Given a single face image, they either optimize~\cite{dib2021practical,thies2016face2face,giebenhain2025pixel3dmm} or learn a neural network to predict 3DMM parameters~\cite{deng2019accurate,DECA:Siggraph2021,wang20243d}.
However, their reconstruction quality is largely constrained by 3DMM itself.
To improve the representation power, some works propose to augment 3DMM by adding per-vertex displacement~\cite{lei2023hierarchical}, predicting a displacement map~\cite{lei2023hierarchical,DECA:Siggraph2021,yang2020facescape,dib2024mosar}, or fine-tuning the 3DMM model on large-scale datasets~\cite{DBLP:conf/cvpr/TewariZ0BKPT18,tewari2019fml,DBLP:conf/cvpr/0001TSET21,han2023ReflectanceMM}.
Although their results are consistent with the 3DMM topology, the reconstruction quality is still confined by the 3DMM prior. 
Another group of works proposes to reconstruct facial geometry without 3DMM~\cite{guo2023rafare,sela2017unrestricted,Zhang_2021_CVPR}.
Among these methods, the most relevant to us is \citet{sela2017unrestricted}; they learn a network to predict a UV coordinate image along with a depth image from the input image and then fit a template mesh to it to ensure topologically consistent reconstruction.
Although we adopt a similar representation, our method works on a multi-view setup.
Thus, we need to fuse the prediction from each view to a consistent mesh, which is challenging but not explored by \citet{sela2017unrestricted}.
In addition, we apply a 3D foundation model, \emph{i.e.} VGGT, to infer the point map instead of using a small model specially trained on facial data. 
Thus, we achieve more robust results as shown in our experiments.

\subsection{Multi-view Face Reconstruction}
From multi-view images, traditional methods typically apply multi-view stereo to reconstruct a detailed scan and then register a template to it~\cite{riviere2020single}, or simultaneously perform these steps~\cite{DBLP:journals/cgf/FyffeNHSBJLD17}.
However, this process is often time-consuming and requires complex manual efforts to ensure high-quality results.
Recent works apply a neural network to predict 3DMM parameters from multi-view images~\cite{bai2020deep,wu2019mvf}.
Even with corrective basis~\cite{bai2020deep}, the construction quality is inevitably limited by the usage of 3DMM.
More recently, some works~\cite{li2021tofu,Bolkart2023Tempeh,li2024grape,DBLP:journals/corr/abs-2201-01016} propose to infer a topologically consistent mesh directly from the multi-view images.
Although high-quality results are demonstrated, they cannot generalize to in-the-wild data captured by daily users, as they are trained on limited 3D facial data captured in the studio.
In this paper, we innovatively apply the point map representation for multi-view topologically consistent facial geometry reconstruction.
Compared to prior arts, our representation not only has powerful representation ability but also inherits strong generalization ability from VGGT.

\subsection{Foundation Models in Face-Related Problems}
By training on large-scale datasets, Vision foundation models\cite{rombach2021highresolution,caron2021emerging,oquab2023dinov2,wang2025vggt} have demonstrated impressive generalization ability on various downstream tasks.
To improve the generalization ability, recent works augment the feature maps with these foundation models during training on face datasets~\cite{giebenhain2025pixel3dmm,kirschstein2025avat3r,ren2025s}.
Instead, we distill high-quality topologically consistent facial geometry from raw predictions of VGGT in a zero-shot manner via test-time optimization.

\section{Method}
Given in-the-wild multi-view images captured around a face by everyday users, our goal is to reconstruct a topologically consistent face mesh from these inputs.
To ensure generalization to in-the-wild capture, we build our method on top of VGGT~\cite{wang2025vggt}, a powerful 3D foundation model.
Specifically, we augment the VGGT raw predictions with facial topology and propose a novel Topology-Aware Bundle Adjustment technique to fuse the prediction in each view to a high-quality topologically consistent mesh.
Before diving into the details, we first introduce the necessary background of VGGT and Bundle Adjustment.

\subsection{Preliminary}
\paragraph{VGGT} 
VGGT~\cite{wang2025vggt} is a recently proposed foundation model for 3D reconstruction.
Given multi-view images $\{I_i\}_{i=1}^{V}, I_i\in\mathbb{R}^{H\times W\times 3}$, it learns a transformer $\mathcal{H}(\cdot)$ to infer the point map and camera intrinsic and extrinsic parameters for each view in a single network pass:
\begin{equation}
\label{eq:method:vggt}
    \{K_i, R_i, t_i, X_i\}_{i=1}^V = \mathcal{H}(\{I_i\}_{i=1}^V)
\end{equation}
Here, $K_i\in\mathbb{R}^{3\times3}$ denotes the camera intrinsic matrix, $R_i\in\mathbb{R}^{3\times3}$ and $t_i\in\mathbb{R}^{3}$ are the camera extrinsic parameters, and $X_i\in\mathbb{R}^{H\times W\times 3}$ is the pixel-aligned point map which records the 3D position of each pixel; the subscript $i$ means the prediction is corresponding to the $i$-th view.
All the predictions of VGGT are in the coordinate frame of the first camera.
Please refer to their paper for more details.

By training on large-scale datasets, VGGT demonstrates superior generalization ability and achieves state-of-the-art results on various benchmarks.
To best maintain its generalization ability, we freeze its pretrained weights and propose novel techniques to distill a topologically consistent mesh from its raw predictions, \emph{i.e.} $\{K_i,R_i,t_i,X_i\}_{i=1}^V$.

\paragraph{Bundle Adjustment}
Bundle Adjustment~\cite{triggs1999bundle} is a process that simultaneously optimizes the initial estimation of the camera parameters $\{K_i, R_i, t_i\}_{i=1}^V$ and $N$ 3D points $\{p_j\}_{j=1}^N$ to make them better match the 2D tracks $q_{ij}$ after the perspective projection $\pi(\cdot)$:
\begin{equation}
\label{eq:method:ba}
    \min_{\{K_i, R_i, t_i\}_{i=1}^V, \{p_j\}_{j=1}^N} \sum_{i=1}^V\sum_{j=1}^N v_{ij}\cdot||\pi(K_i,R_i,t_i,p_j)-q_{ij}||^2_2
\end{equation}
Here, $p_j\in\mathbb{R}^3$ is the 3D point, $q_{ij}\in\mathbb{R}^2$ is the target screen coordinate of $p_j$ on the $i$-th view, and $v_{ij}\in\{0,1\}$ is a flag indicating whether the track is valid.
Note we use the subscript $i$ to index the view and $j$ to index the 3D points; we will also adhere to this convention in the following paper.
Eq.~\eqref{eq:method:ba} is a non-linear least square problem, which is well-studied in the literature and can be solved efficiently by the Ceres solver~\cite{Agarwal_Ceres_Solver_2022}.

\begin{figure*}[t]
    \centering
    \includegraphics[width=1.\textwidth]{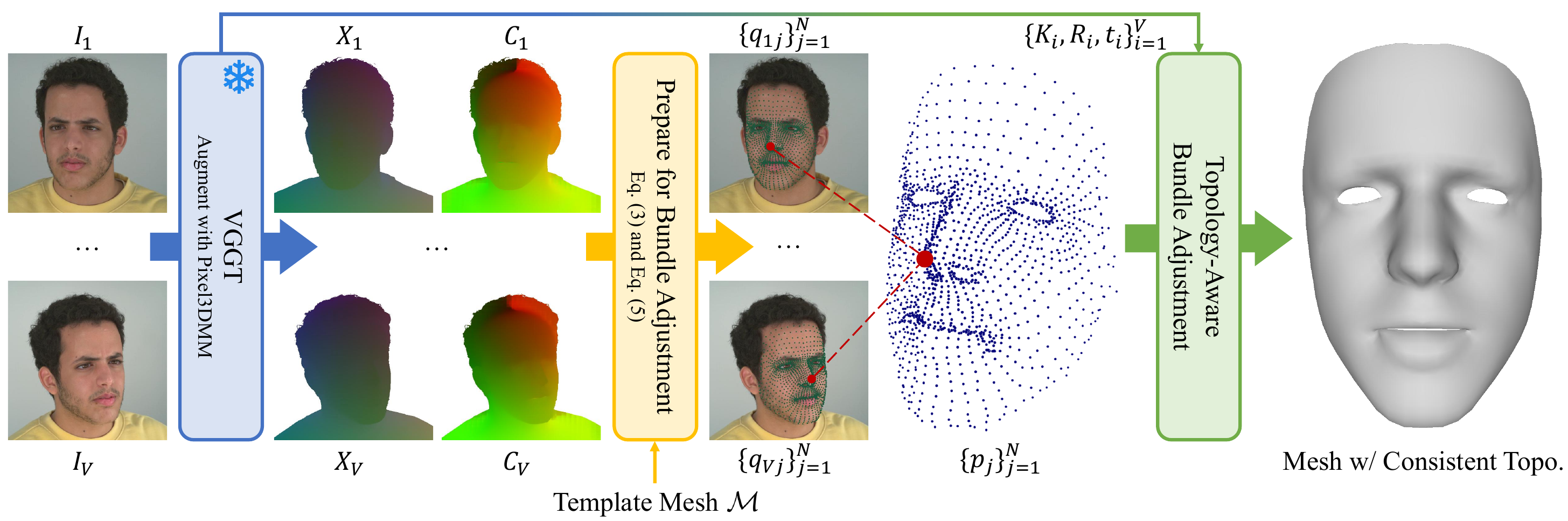}
    \caption{
        The framework of VGGTFace. Given $V$ multi-view images $\{I_i\}_{i=1}^V$ as input, we first augment VGGT with Pixel3DMM to obtain the camera parameters $\{K_i,R_i,t_i\}_{i=1}^V$, point map $\{X_i\}_{i=1}^V$, and UV coordinate image $\{C_i\}_{i=1}^V$ for each view. We then convert these raw predictions to a set of tracks $\{q_{ij}\}$ and an initial point cloud $\{p_j\}_{j=1}^N$ according to the template mesh $\mathcal{M}$. Next, we propose a novel Topology-Aware Bundle Adjustment technique to simultaneously optimize the camera parameters $\{K_i,R_i,t_i\}_{i=1}^V$ and the point cloud $\{p_j\}_{j=1}^N$ to better match the tracks.
        After that, we connect the point cloud with the topology of $\mathcal{M}$ to obtain a mesh with consistent topology.
    }
    \label{Fig:framework}
\end{figure*}

\subsection{Augment VGGT Predictions with Topology}
As shown in Figure~\ref{Fig:framework}, given $V$ multi-view face images captured by everyday users, we first feed them to VGGT to obtain the point map $X_i$ and camera parameters $\{K_i,R_i,t_i\}$ for each view using Eq.~\eqref{eq:method:vggt}.

However, applying VGGT for topologically consistent facial geometry reconstruction is not easy, as the raw prediction of VGGT does not contain any topology information, as shown in Eq.~\eqref{eq:method:vggt}.
Our key observation is that the geometry of VGGT is represented in a pixel-aligned fashion, \emph{i.e.} the point map $X_i$.
Thus, we can inject extra information into the prediction of VGGT using any screen-space method, which is a format widely adopted in face-related problems such as face parsing~\cite{zheng2022farl} and facial intrinsic decomposition~\cite{kim2024switchlight}.

To augment the predictions of VGGT with topology, we use an off-the-shelf method, Pixel3DMM~\cite{giebenhain2025pixel3dmm}, to predict the UV coordinate image $\{C_i\}_{i=1}^V, C_i\in \mathbb{R}^{H\times W\times 2}$ for each view as shown in Figure~\ref{Fig:framework}.
This way, for each pixel in each view, we have not only its 3D position predicted by VGGT, but also its UV coordinate predicted by Pixel3DMM.
Then, we can use the UV coordinate to build correspondence between the vertex on the template mesh and the pixel in each view. 
After that, we can transfer the topology information defined on the template mesh to the VGGT's prediction.

Specifically, given a template mesh $\mathcal{M}$\footnote{$\mathcal{M}$ should have the same UV definition as the UV predictor, \emph{i.e.} Pixel3DMM.} that has $N$ vertices with the corresponding UV coordinate $\{c_j\}_{j=1}^N,c_j\in\mathbb{R}^2$ and the topology $\mathcal{F}$ which defines the connection information of the $N$ vertices, we can lookup its corresponding pixel on the $i$-th view via nearest neighbor search in the UV distance space:
\begin{equation}
\label{eq:method:knn}
    \{q_{ij}^x,q_{ij}^y\} = \arg\!\min ||C_i[q_{ij}^x][q_{ij}^y] - c_j||_2
\end{equation}
We denote $q_{ij}=(q_{ij}^x,q_{ij}^y),q_{ij}\in\mathbb{R}^2$ as the corresponding screen coordinate of the $j$-th template vertex on the $i$-th view.
Note that not all the vertices are visible in each view; we realize this visibility using a flag $v_{ij}$ by thresholding the nearest distance with a tolerance $\epsilon$:
\begin{equation}
\label{eq:method:vis}
\begin{aligned}
    v_{ij} = \begin{cases} 
    0 & ||C_i[q_{ij}^x][q_{ij}^y] - c_j||_2\geq \epsilon \\
    1 & ||C_i[q_{ij}^x][q_{ij}^y] - c_j||_2<\epsilon
    \end{cases}
\end{aligned}
\end{equation}
Then, we can transfer the topology $\mathcal{F}$ to the VGGT's predicted 3D point $\{X_i[q_{ij}^x][q_{ij}^y]\}_{j=1}^N$ on the $i$-th view. 

However, as VGGT are not perfectly accurate, the 3D position of the same template vertex computed from different views are often inconsistent, \emph{i.e.} $X_{i_1}[q_{i_1j}^x][q_{i_1j}^y]\neq X_{i_2}[q_{i_2j}^x][q_{i_2j}^y]$ where $i_1$ and $i_2$ are the index of two different views.
Naively fusing the 3D position computed from different views does not guarantee valid results, as shown in Figure~\ref{Fig:ablat} (the \emph{w/o BA} baseline). 
That is because the fusion results are not supervised during the training of VGGT.
To solve this problem, our key insight is to regard the screen coordinates of the template vertex, \emph{i.e.} $q_{ij}$, as tracks and formulate the point cloud fusion as the Bundle Adjustment problem.

\subsection{Topology-Aware Bundle Adjustment}
To enable Bundle Adjustment, we first fuse the same template vertex computed from different views, \emph{i.e.} $\{X_i[q_{ij}^x][q_{ij}^y]\}_{i=1}^V$, to a single 3D point $p_j$ to serve as the initial point cloud of Eq.~\eqref{eq:method:ba}; recall that $X_i[q_{ij}^x][q_{ij}^y]$ is the 3D position of the $j$-th template vertex computed from the $i$-th view.
We define the fusing operator as a simple weighted average by the visibility $v_{ij}$; this is not a critical step and can be implemented in other ways, as the fused point cloud only acts as an initialization in BA:
\begin{equation}
\label{eq:method:fuse}
    p_j = \frac{\sum_{i=1}^V v_{ij}\cdot X_i[q_{ij}^x][q_{ij}^y]}{\sum_{i=1}^V v_{ij}}
\end{equation}
We visualize $\{p_j\}_{j=1}^N$ and the corresponding tracks $q_{ij}$ in Figure~\ref{Fig:framework}.
In practice, some template vertices are invisible to all the views because they cannot pass the visibility test in Eq.~\eqref{eq:method:vis}, \emph{i.e.} for some $j$, $\sum_{i=1}^V v_{ij}$ is zero; for these points, we simply set their fused position to be the origin.

Now, by augmenting with Pixel3DMM, we have transferred the raw prediction of VGGT to the combination of per-view camera parameters $\{K_i,R_i,t_i\}$, a set of tracks $q_{ij}$ which denotes the screen coordinates of the $j$-th template vertex on the $i$-th view along with their visibility $v_{ij}$, and a point cloud $\{p_j\}_{j=1}^N$ which denotes the initial 3D positions of the template vertices.
Then, we can directly send these parameters to Eq.~\eqref{eq:method:ba} to optimize the camera parameters $\{K_i,R_i,t_i\}$ and the position of the template vertices $\{p_j\}_{j=1}^N$ to best match the tracks $q_{ij}$.
After the Bundle Adjustment, we can connect $\{p_j\}_{j=1}^N$ via the topology $\mathcal{F}$ to obtain a topologically consistent mesh.

However, we observe that the resulting mesh is noisy if naively following the optimization in Eq.~\eqref{eq:method:ba}, as shown in Figure~\ref{Fig:ablat} (the \emph{w/o Laplacian} baseline).
That is because the tracks $q_{ij}$ are computed from Pixel3DMM, which is a neural network and thus not perfectly accurate. 
In addition, if a template vertex has only one or zero visible tracks, \emph{i.e.} $\sum_{i=1}^V v_{ij}<2$, it is unconstrained in Eq.~\eqref{eq:method:ba}~\cite{triggs1999bundle}.
To address the inaccurate track problem, traditional works like COLMAP~\cite{schoenberger2016mvs,schoenberger2016sfm} typically require tedious parameter tweaking to filter out the influence of low-quality tracks.
Although more robust to the inaccurate tracks, they still cannot handle points with fewer than two visible tracks.
Ignoring these points without enough visible tracks is unacceptable in our method, as all the points correspond to the template mesh; deleting any of them will destroy the topology.

Fortunately, our key insight is that our 3D points $\{p_j\}_{j=1}^N$ have topology, which is not satisfied in the scenario of existing works.
Motivated by this observation, we propose adding a Laplacian term to the Bundle Adjustment objective:
\begin{multline}
\label{eq:method:toba}
    \min_{\{K_i, R_i, t_i\}_{i=1}^V, \{p_j\}_{j=1}^N} \sum_{i=1}^V\sum_{j=1}^N v_{ij}\cdot||\pi(K_i,R_i,t_i,p_j)-q_{ij}||^2_2 \\
    + \sum_{j=1}^N ||p_j - \frac{1}{|\mathcal{N}(j)|}\cdot\sum_{k\in\mathcal{N}(j)}p_k||
\end{multline}
Here, $\mathcal{N}(j)$ denotes the 1-ring neighborhood list of the $j$-th vertex on the template mesh $\mathcal{M}$, which can be computed efficiently given the topology $\mathcal{F}$, and $|\mathcal{N}(j)|$ denotes the number of neighborhoods of the $j$-th vertex. 
This Laplacian term not only enforces a smooth regularization on the 3D points to make the optimization more robust to inaccurate tracks, but also adds a direct constraint on the 3D points that lack enough visible tracks.
We emphasize that the latter benefit, \emph{i.e.}, ensuring every 3D point is supervised in the BA process, is important in our problem, but has not been explored in the literature.
As shown in Figure~\ref{Fig:ablat}, our Topology-Aware Bundle Adjustment method obtains significantly better results than the naive baseline.

In practice, we modify the original BA Ceres solver to solve Eq.~\eqref{eq:method:toba}.
Specifically, we iteratively minimize the original re-projection error and the Laplacian constraint until convergence.
Thanks to the highly optimized Ceres solver, the whole method takes less than 10 seconds to process 16 views on a single NVIDIA RTX 4090, including the inference time of VGGT and Pixel3DMM, the time to prepare inputs for BA, and the optimization process of Eq.~\eqref{eq:method:toba}.

\section{Experiments}
We first provide implementation details of our method.
We then compare our method to baseline methods on various benchmarks.
Next, we evaluate several key design choices in our method.
Lastly, we discuss the limitations and future works.
Due to the page limit, we urge the reader to check our \emph{suppl. material} for more results.
\subsection{Implementation Details}
Our system takes 16 input views unless otherwise specified. Although it can process arbitrary numbers of images, we find 16 provides a good balance between view coverage and runtime; for in-the-wild videos, we uniformly sample 16 frames. Input images are center-cropped and resized to $518{\times}518$ before being fed to VGGT. Following \citet{wang2025vggt}, we compute the point map using VGGT's depth-branch output, which is empirically more accurate than its point-map head. The visibility threshold $\epsilon$ in Eq.~\eqref{eq:method:vis} is set as the $70$-th percentile of the nearest-distance distribution in each view, which performs slightly better than using a fixed value. We adopt FLAME~\cite{FLAME:SiggraphAsia2017} as the template mesh $\mathcal{M}$ for its UV compatibility with Pixel3DMM. Importantly, FLAME is used only for topology (vertex connectivity and UV layout); no FLAME shape or expression bases are used. The facial region is subdivided to approximately $5000$ vertices. All experiments are performed on an NVIDIA RTX~4090.
\begin{figure*}[t]
    \centering
    \includegraphics[width=1.\textwidth]{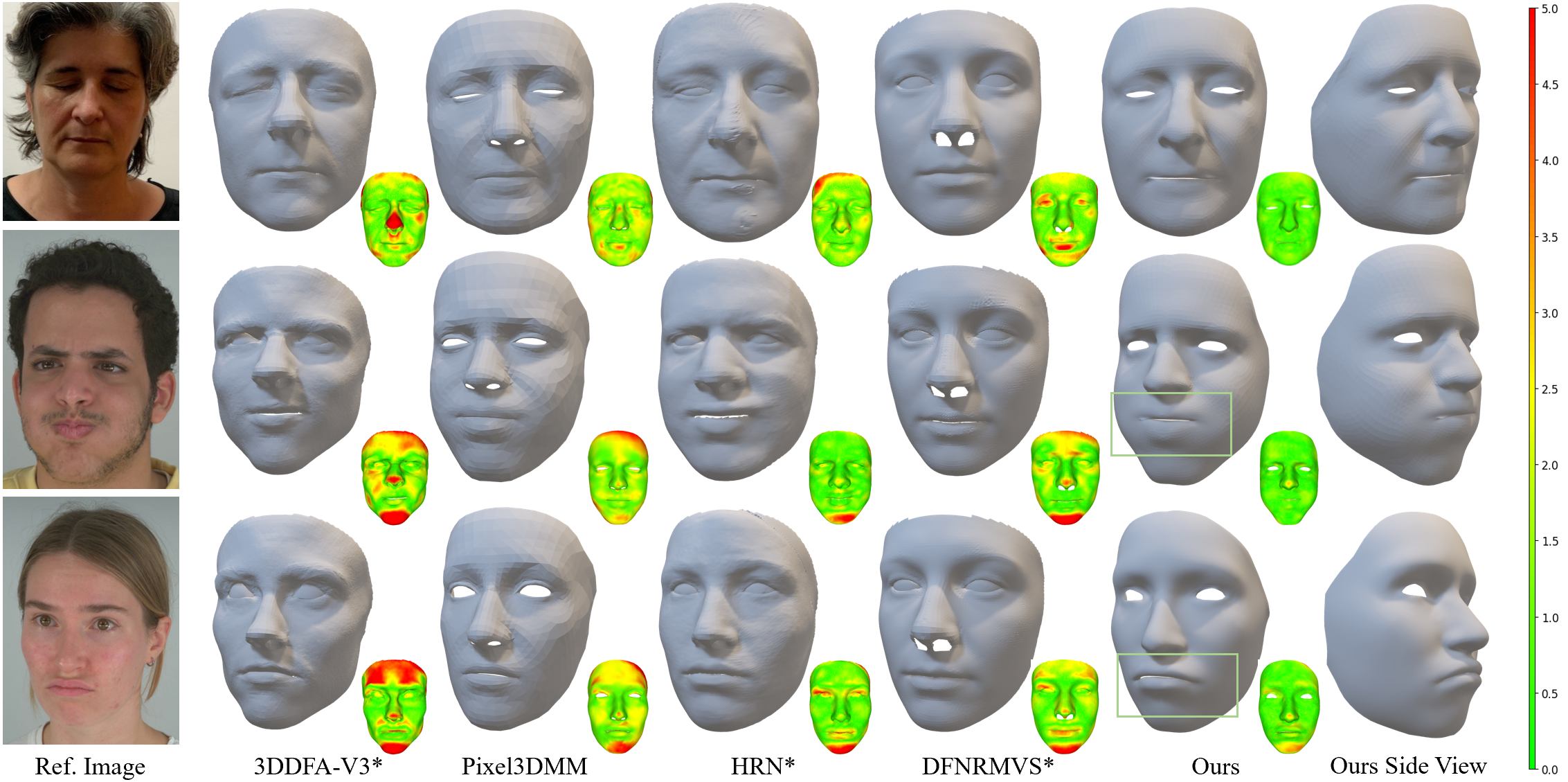}
    \caption{
        Qualitative comparison on the H3DS benchmark (the first row) and the NeRSemble benchmark (the last two rows). For each result, we pair it with the error map against the GT scan. Methods marked with * adopt the BFM topology with around 40000 vertices, while Ours and Pixel3DMM follow the FLAME topology with around 5000 and 1500 vertices, respectively.
    }
    \label{Fig:cmp}
\end{figure*}

\begin{table}[t]
\renewcommand{\arraystretch}{0.9}
\setlength{\tabcolsep}{3pt}
\centering
\resizebox{\columnwidth}{!}{%
\begin{tabular}{@{}l c c c c c c@{}}
\toprule
\multirow{2}{*}{Method} & 
\multicolumn{1}{c}{H3DS} & 
\multicolumn{3}{c}{NeRSemble} & 
\multirow{2}{*}{Time} &
\multirow{2}{*}{\# Views} \\
\cmidrule(lr){2-2} \cmidrule(lr){3-5}
& Mean $\downarrow$ & Mean $\downarrow$ & Median $\downarrow$ & Std $\downarrow$ & & \\
\midrule
DECA        & 1.99 & 1.58 & 1.55 & 1.18 & $< 1s$ &  1 \\
3DDFA-V3    & 1.81 & 1.52 & 1.46 & 1.08 & $< 1s$ &  1 \\
Pixel3DMM   & 1.69 & 1.42 & 1.38 & 1.01 & $\sim 70s$ &  1 \\
DFNRMVS     & 1.78 & 1.62 & 1.58 & 1.09 & $\sim 30s$ &  16 \\
HRN         & 1.46 & 1.49 & 1.43 & 1.10 & $\sim 160s$ &  16 \\
NeuS2       & 2.11 & 2.58 & 2.52 & 2.39 & $\sim 300s$ &  16 \\
\midrule
Ours        & \textbf{1.18} & \textbf{0.98} & \textbf{0.98} & \textbf{0.65} & $< 10s$ &  16 \\
\midrule
w/o BA & 1.52 & 1.44 & 1.37 & 1.04 & $< 10s$ & 16 \\
w/o Laplacian &  1.20 & 0.99 & 0.98 & 0.65 & $< 10s$ & 16 \\
\bottomrule
\end{tabular}
}
\caption{Quantitative comparison on H3DS and NeRSemble benchmark. All metrics are in millimeters (mm).}
\label{tab:comparison}
\end{table}

\subsection{Comparisons}
We first introduce the baseline methods we used for comparison.
Then, we report the comparison results on the H3DS~\cite{ramon2021h3d,caselles2025implicit} and NeRSemble benchmark~\cite{kirschstein2023nersemble}.

\paragraph{Baseline Methods}
We compare state-of-the-art methods that can reconstruct a topologically consistent mesh from single-view (DECA~\cite{DECA:Siggraph2021}, 3DDFA-V3~\cite{wang20243d}, and Pixel3DMM~\cite{giebenhain2025pixel3dmm}) or multi-view face images (HRN~\cite{lei2023hierarchical} and DFNRMVS~\cite{bai2020deep}).
We also involve \citet{sela2017unrestricted} for comparison in the \emph{suppl. material}, as their method adopts a similar representation as our method, but in a single-view setup.
In addition, we compare to NeuS2~\cite{neus2}, a well-known multi-view geometry reconstruction method for generic objects.
We do not compare to ToFu~\cite{li2021tofu} and TEMPEH~\cite{Bolkart2023Tempeh} as their method cannot generalize to camera configurations different from their training studio.
Although GRAPE~\cite{li2024grape} claims it can process camera configurations different from their training studio, we do not compare to it as its code is not released.

\paragraph{Comparison on H3DS Benchmark}
The H3DS benchmark~\cite{ramon2021h3d,caselles2025implicit} contains 59 subjects with neutral expression.
Following SIRA++~\cite{caselles2025implicit}, we report the mean chamfer distance \emph{w.r.t} to the ground truth mesh in Table~\ref{tab:comparison} and present qualitative comparisons in Figure~\ref{Fig:cmp}.
Compared to existing works, our method obtains the best results.
In addition, as an optimization-based method, ours is still significantly faster than other optimization-based methods such as Pixel3DMM, HRN, DFNRMVS, and NeuS2. 
That is because our method is based on the highly optimized BA solver.

\paragraph{Comparison on NeRSemble Benchmark}
As the H3DS benchmark only contains subjects with neutral expressions, we conduct comparisons on the more challenging NeRSemble dataset~\cite{kirschstein2023nersemble} with facial expression variation.
The original NeRSemble dataset only contains 16-view videos of multiple subjects performing expression sequences.
We build a benchmark for facial geometry evaluation on top of it\footnote{Although Pixel3DMM also builds a similar geometry reconstruction benchmark on NeRSemble, upon the submission of our paper, their benchmark is not publicly available.}.
Specifically, we manually select 76 frames from the data of 40 subjects; see more details in our \emph{suppl. material}.
For each subject, we use MetaShape\footnote{https://www.agisoftmetashape.com} to reconstruct a mesh from the provided 16-view images, which serves as the ground truth facial geometry (GT scan).
We then manually annotate 6 landmarks on the GT scan for alignment.
Given a reconstructed face mesh and the corresponding 6 landmarks, we first rigidly align it to the GT scan, and then compute the mean/median/std of the chamfer distance \emph{w.r.t} the GT scan.

As shown in Table~\ref{tab:comparison}, our method is the only one that achieves an error less than 1 millimeter.
As shown in Figure~\ref{Fig:cmp}, our method can reconstruct challenging expressions with high fidelity.
Compared to single-view methods, \emph{e.g.} 3DDFA-V3 and Pixel3DMM, our method can better reconstruct the mouth shape due to the exploitation of multi-view information, which inherently has ambiguity in the single-view setup.
Compared to multi-view methods, \emph{e.g.} HRN and DFNRMVS, our method has stronger representation power stemming from the point map representation to model person-specific facial characteristics such as puffing and pouting (highlighted with green boxes in Figure~\ref{Fig:cmp}); despite per-vertex displacement or corrective basis being adopted, HRN and DFNRMVS are inevitably constrained by their dependency on the 3DMM.

\begin{figure}[t]
    \centering
    \includegraphics[width=0.45\textwidth]{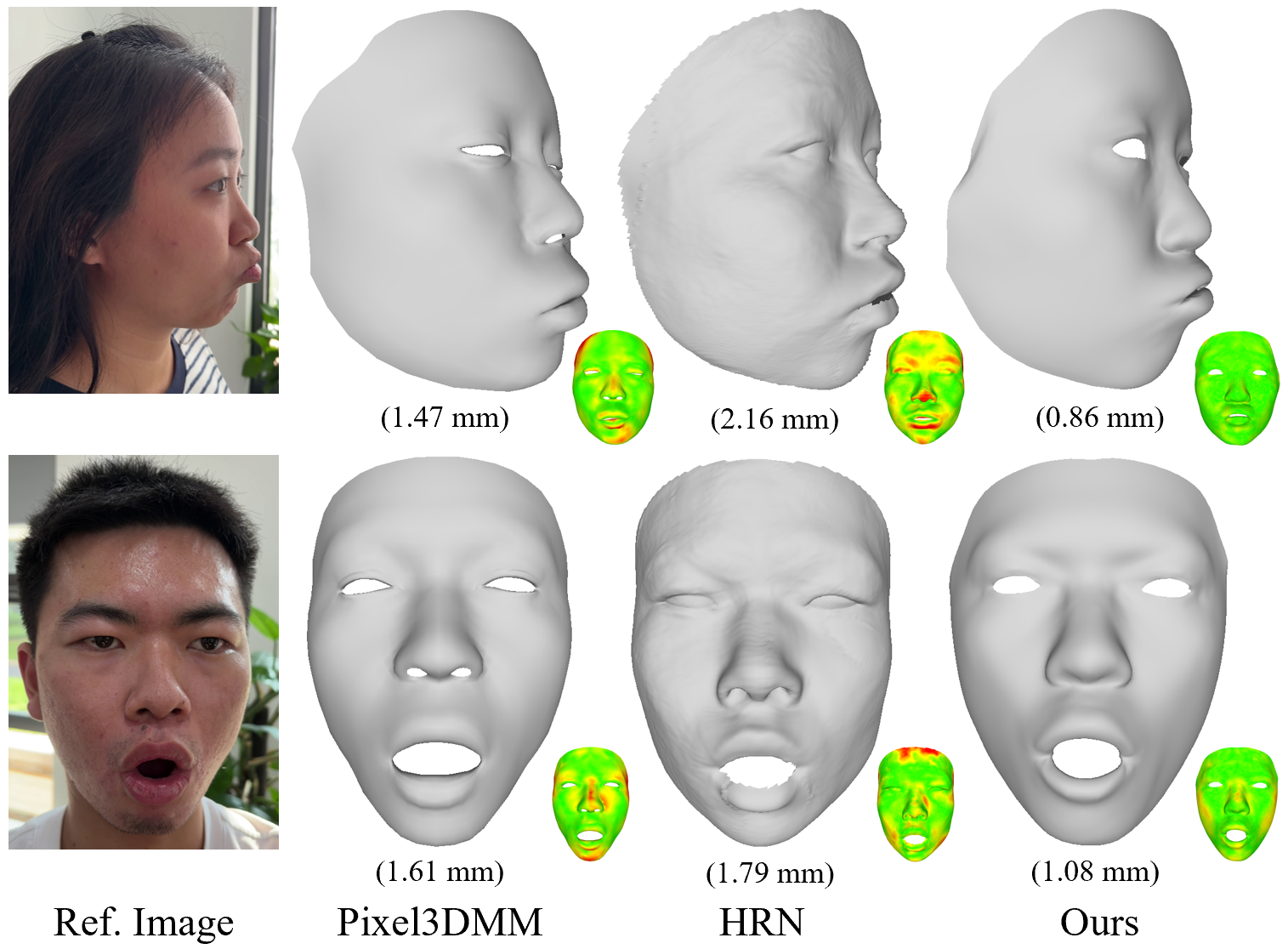}
    \caption{
        Comparisons on in-the-wild data. For each result, we pair it with the error map against the GT scan with the same colorbar as Figure~\ref{Fig:cmp}. We report the mean chamfer distance below each method's result.
    }
    \label{Fig:cmp_wild}
\end{figure}

\paragraph{Comparison on In-the-Wild Data}
As shown in Figure~\ref{Fig:cmp_wild}, we conduct qualitative and quantitative comparisons on in-the-wild data captured by ourselves.
We involve Pixel3DMM and HRN here, as these two methods perform better in the H3DS and NeRSemble benchmark.
We capture a smartphone video around the subject and sample 16 frames as the inputs to our method.
We run MetaShape on all the frames to reconstruct a mesh as the GT scan.
We then align the mesh to compute the chamfer distance the same way as we do in building the NeRSemble benchmark.
Again, our method obtains the best mean chamfer distance.

\begin{figure}[t]
    \centering
    \includegraphics[width=0.45\textwidth]{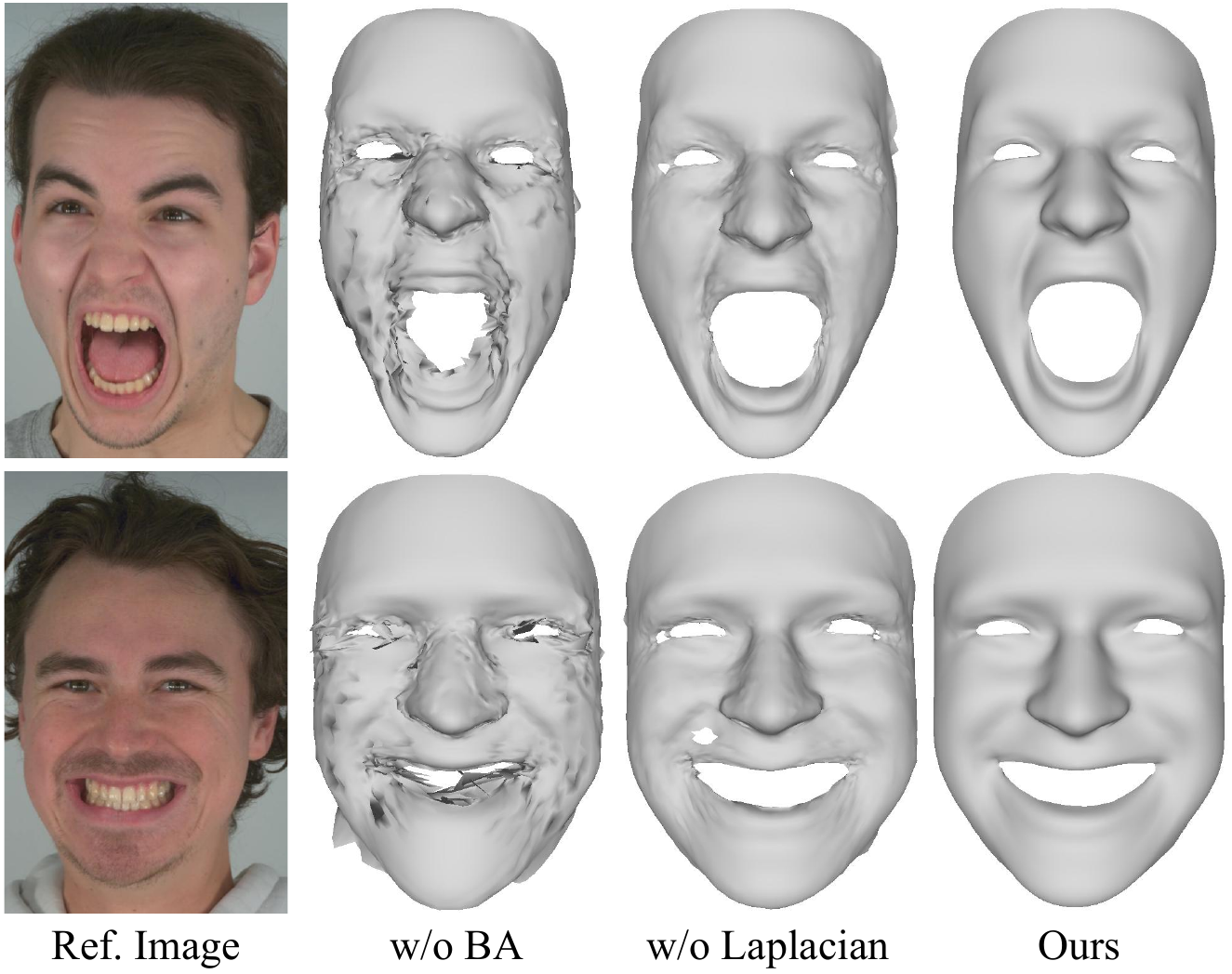}
    \caption{
        Qualitative evaluation of the key design choices in our method.
    }
    \label{Fig:ablat}
\end{figure}

\subsection{Evaluations}
We first evaluate the effectiveness of applying Bundle Adjustment to fuse the initial point cloud estimated by VGGT. 
For the \emph{w/o BA} baseline, we directly connect the points in Eq.~\eqref{eq:method:fuse} according to the template topology $\mathcal{F}$. 
As shown in Figure~\ref{Fig:ablat}, this naive fusion leads to noticeable surface artifacts, as the predictions of VGGT and Pixel3DMM are not fully accurate and the fused result is not supervised during their training. 
The quantitative results in Table~\ref{tab:comparison} also verify the benefit of BA.

We then evaluate the role of the Laplacian term in the BA objective. 
The \emph{w/o Laplacian} baseline applies the naive BA in Eq.~\eqref{eq:method:ba}. 
Although it improves upon \emph{w/o BA}, the reconstructed surfaces remain noisy due to imperfect correspondences, and some vertices disappear when fewer than two valid tracks are available. 
With the proposed TopBA, the Laplacian energy enforces local consistency and results in a clean and complete mesh. 
Again, Table~\ref{tab:comparison} confirms the superiority of our full method.

\begin{figure}[t]
    \centering
    \includegraphics[width=0.45\textwidth]{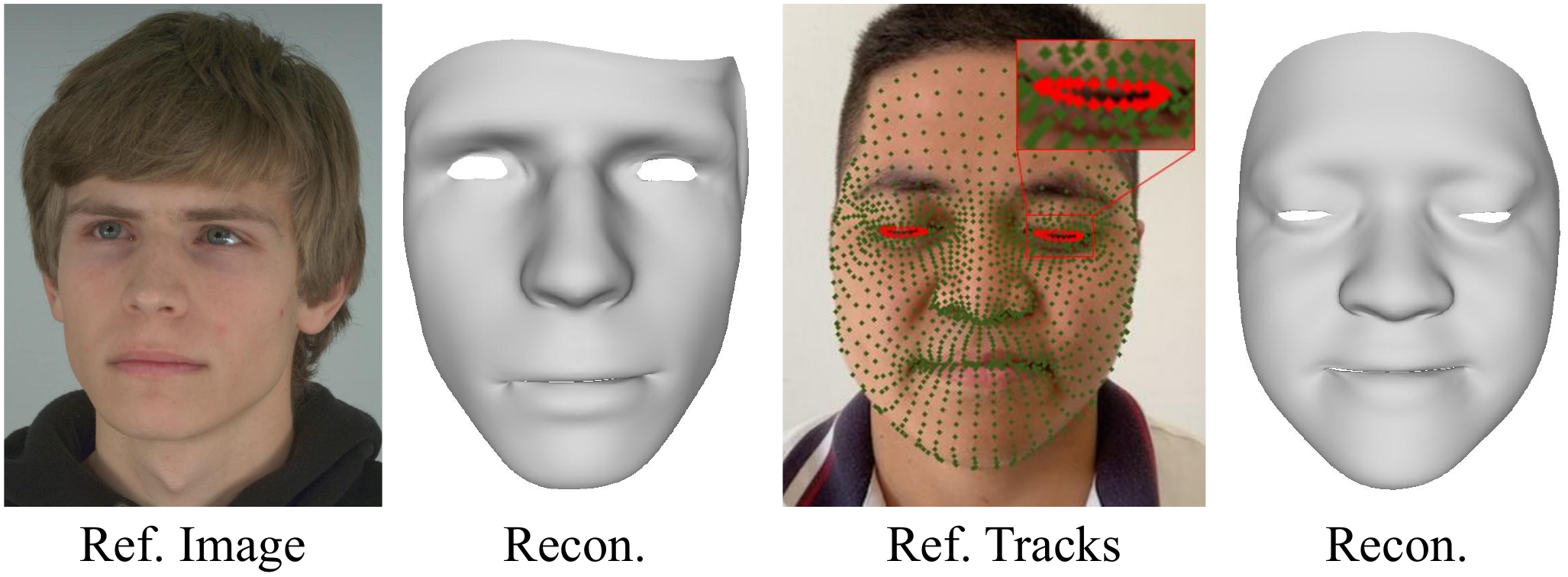}
    \caption{
        Limitations of our method. 
    }
    \label{Fig:limit}
\end{figure}

\subsection{Limitations and Discussions}
First, as a mesh-based method, our method has limitations in representing hair strands, as shown in Figure~\ref{Fig:limit}.
Secondly, constrained by the inaccuracy of the tracks predicted by Pixel3DMM (visualized in the Ref. Tracks image in Figure~\ref{Fig:limit}), we observe inaccurate reconstruction results on subjects with eyes closed, as shown in Figure~\ref{Fig:limit}.
Incorporating special losses for eye regions is our future work~\cite{wang20243d,giebenhain2025pixel3dmm}.
Besides, our method focuses on reconstructing a large-scale static facial geometry without texture.
Predicting a displacement or normal map on top of it~\cite{li2021tofu} is an interesting direction, but beyond the scope of this paper.
Combining existing works to build a complete avatar with blendshapes~\cite{ming2024high}, and reflectance maps~\cite{han2024cora,han2025dora} is an interesting future direction.

\section{Conclusion}
We propose VGGTFace, a novel method for topologically consistent facial geometry reconstruction from in-the-wild multi-view face images captured by everyday users.
The core idea is to apply the recently-proposed 3D foundation model, \emph{i.e.} VGGT, for face reconstruction.
As the prediction of VGGT does not contain topology information, we propose to augment it with pixel-aligned UV coordinate images predicted by Pixel3DMM. 
To fuse the VGGT point cloud from different views, we propose a novel Topology-Aware Bundle Adjustment strategy, where we construct a Laplacian energy to the Bundle Adjustment objective, making our method robust to the inaccurate tracks.
Experiments demonstrate our method obtains state-of-the-art results on various benchmarks and strong generalization ability on in-the-wild data captured by everyday users.

\section*{Acknowledgments}
This work was supported by the National Key R\&D Program of China (2023YFC3305600). This work was also supported by THUIBCS, Tsinghua University, and BLBCI, Beijing Municipal Education Commission.

\bibliography{aaai2026}

\clearpage            %

\appendix

\begin{center}
{\LARGE\bfseries
VGGTFace: Topologically Consistent Facial Geometry Reconstruction in the Wild\\[4pt]
\emph{Supplementary Material}
\par}
\vspace{1em}
\end{center}

\begin{figure*}[ht]
    \centering
    \includegraphics[width=\textwidth]{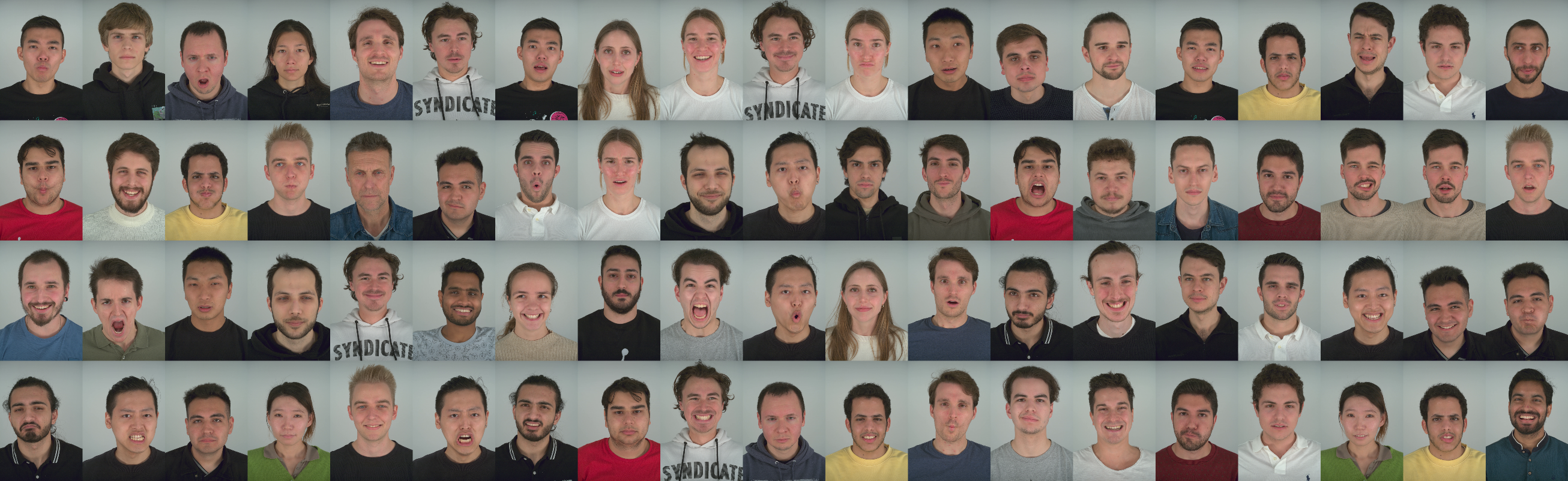}
    \caption{
        Frontal view images of the selected frames in our NeRSemble benchmark~\cite{kirschstein2023nersemble}.
    }
    \label{Fig:ners}
    \vspace{-1em}
\end{figure*}

\section{More Details on the NeRSemble Benchmark}
In Figure~\ref{Fig:ners}, we show the frontal view of the 76 selected frames from 40 subjects in the original NeRSemble dataset.
Our benchmark features diverse and challenging facial expressions, such as puffing and pouting.

\section{More Details on Our Captured Data}

\begin{figure}[t]
    \centering
    \includegraphics[width=0.475\textwidth]{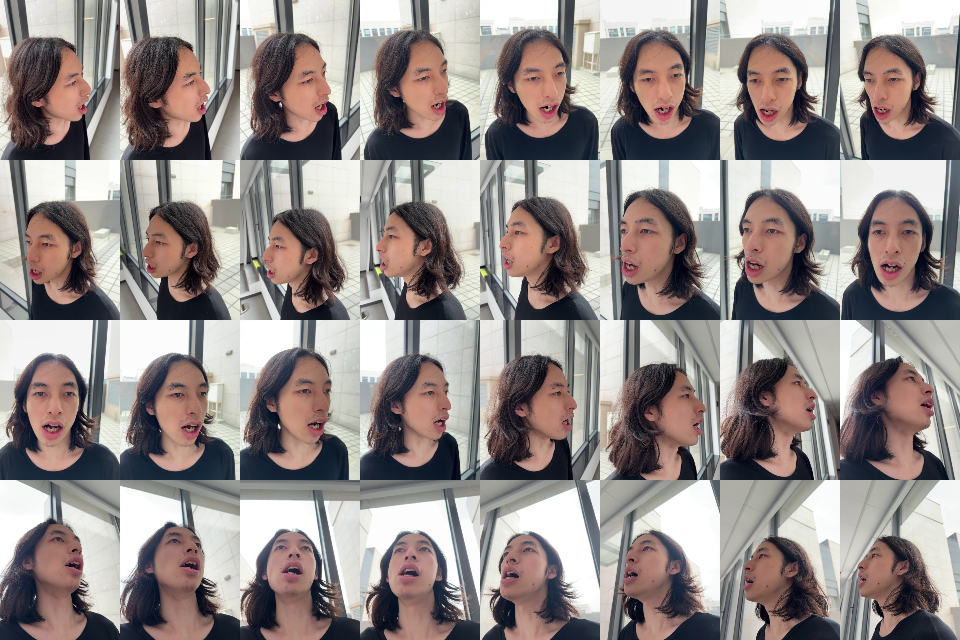}
    \caption{
        Montage of densely sampled frames from one of our self-captured in-the-wild videos.
    }
    \label{Fig:grid}
\end{figure}

We provide additional details about our self-captured in-the-wild dataset. 
All videos are recorded using a hand-held mobile phone, and we have conducted experiments with both iPhone and Android devices (1920$\times$1080 resolution). 
For each subject, we capture a video while moving around the head at a distance of roughly $0.8\sim1.5$ meters. 
From each video, we uniformly sample 16 frames, which serve as the multi-view input to our system. 
Figure~\ref{Fig:grid} shows a montage of densely sampled frames from one captured video, illustrating the overall capture trajectory.

\begin{figure}[t]
    \centering
    \includegraphics[width=0.475\textwidth]{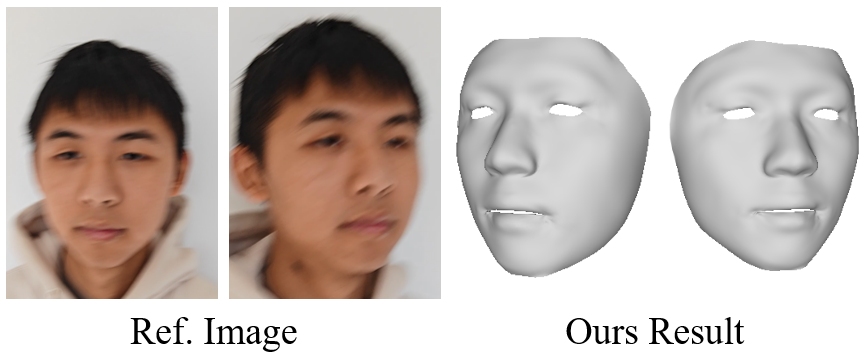}
    \caption{
        Our method remains robust under motion blur. Even when the input images contain noticeable blur (left), the reconstructed mesh (right) preserves accurate facial geometry..
    }
    \label{Fig:blur}
\end{figure}

Our captured data cover a variety of real-world conditions, including mixed and asymmetric lighting, shadows, specular highlights, and natural facial expression changes. 
We empirically find that our method remains robust under practical capture imperfections such as motion blur and slight hand shake.  
Figure~\ref{Fig:blur} presents reconstruction results under noticeable motion blur, where our method still produces high-quality geometry.

\section{More Experiments}
We highly urge the reader to check our \emph{suppl. video} for more comparison results on the NeRSemble benchmark and in-the-wild data captured by ourselves.

\begin{figure}[t]
    \centering
    \includegraphics[width=0.475\textwidth]{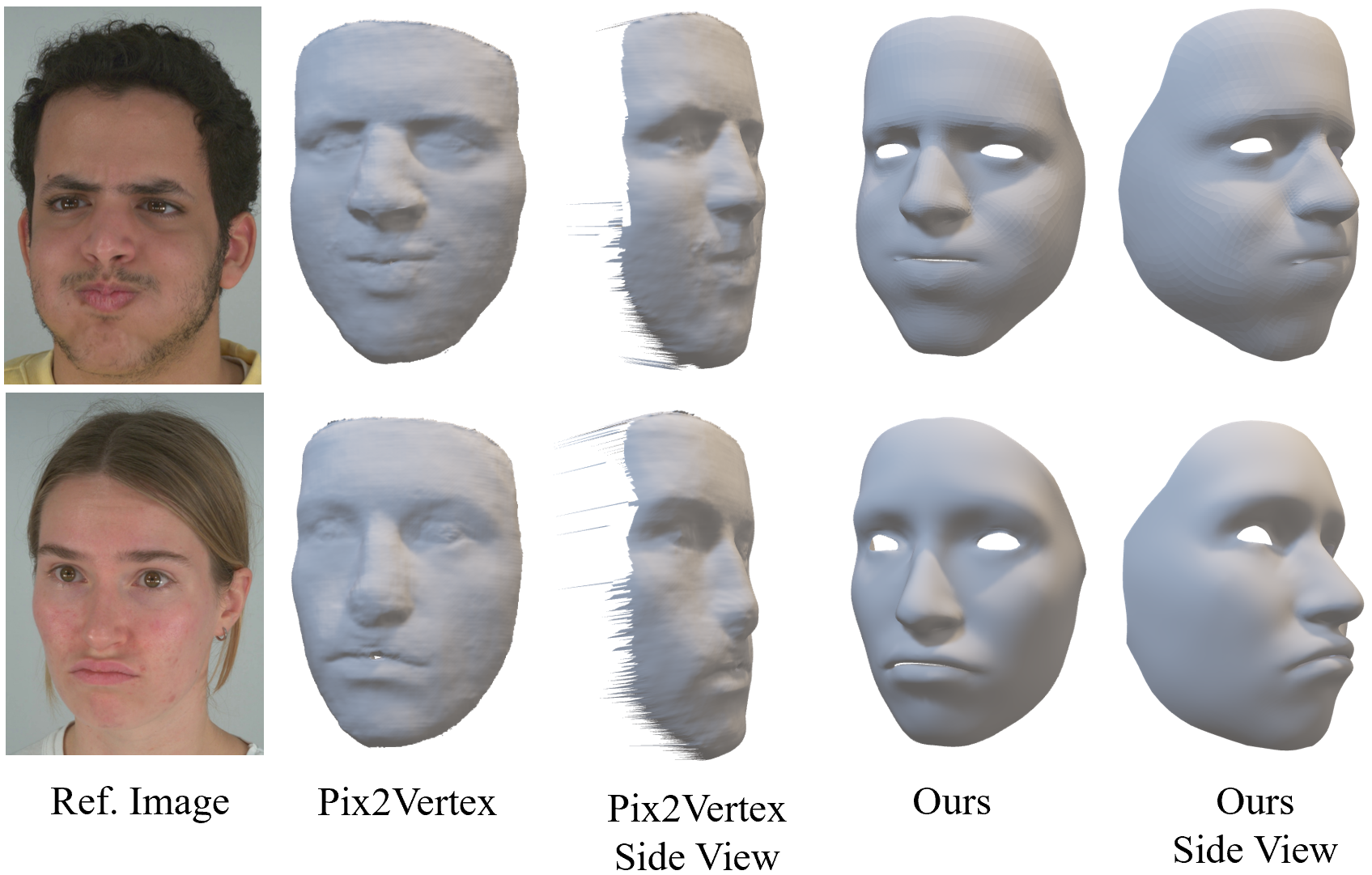}
    \caption{
        Qualitative comparison to Pix2Vertex.
    }
    \label{Fig:p2v}
    \vspace{-0.5em}
\end{figure}

\subsection{Comparison to Pix2Vertex}
We qualitatively compare to Pix2Vertex~\cite{sela2017unrestricted} in Figure~\ref{Fig:p2v}.
The released code of Pix2Vertex\footnote{https://github.com/eladrich/pix2vertex.pytorch} does not contain its proposed template-fitting stage.
Thus, the results of Pix2Vertex only contain a screen-space geometry.
As shown in Figure~\ref{Fig:p2v}, our method yields better results than Pix2Vertex, with a more accurate facial shape (as illustrated more clearly in the side view), as it fully leverages the capabilities of VGGT~\cite{wang2025vggt} and multi-view information. 
Although Pix2Vertex adopts a similar facial geometry representation as our method, it is a small model trained on limited facial data with inferior generalization ability and confined to its single-view setup. 

\begin{table}[t]
\renewcommand{\arraystretch}{0.9}
\setlength{\tabcolsep}{3pt}
\centering
\resizebox{\columnwidth}{!}{%
\begin{tabular}{@{}l c c c c c c@{}}
\toprule
\multirow{2}{*}{Method} & 
\multicolumn{1}{c}{H3DS} & 
\multicolumn{3}{c}{NeRSemble} & 
\multirow{2}{*}{Time} &
\multirow{2}{*}{\# Views} \\
\cmidrule(lr){2-2} \cmidrule(lr){3-5}
& Mean $\downarrow$ & Mean $\downarrow$ & Median $\downarrow$ & Std $\downarrow$ & & \\
\midrule
DECA        & 1.99 & 1.58 & 1.55 & 1.18 & $< 1s$ &  1 \\
3DDFA-V3    & 1.81 & 1.52 & 1.46 & 1.08 & $< 1s$ &  1 \\
Pixel3DMM   & 1.69 & 1.42 & 1.38 & 1.01 & $\sim 70s$ &  1 \\
DFNRMVS     & 1.78 & 1.62 & 1.58 & 1.09 & $\sim 30s$ &  16 \\
HRN         & 1.46 & 1.49 & 1.43 & 1.10 & $\sim 160s$ &  16 \\
NeuS2       & 2.11 & 2.58 & 2.52 & 2.39 & $\sim 300s$ &  16 \\
\midrule
Ours        & \textbf{1.18} & \textbf{0.98} & \textbf{0.98} & \textbf{0.65} & $< 10s$ &  16 \\
\midrule
w/o BA & 1.52 & 1.44 & 1.37 & 1.04 & $< 10s$ & 16 \\
w/o Laplacian &  1.20 & 0.99 & 0.98 & 0.65 & $< 10s$ & 16 \\
naive FLAME & 1.48  & 1.45  & 1.38  & 1.02  & $\sim 15s$  & 16 \\
\bottomrule
\end{tabular}
}
\caption{Quantitative comparison on H3DS and NeRSemble benchmark. All metrics are in millimeters (mm).}
\label{tab:comparison}
\end{table}

\subsection{Comparison to FLAME fitting}
We conduct a naive baseline that combines VGGT~\cite{wang2025vggt}, Pixel3DMM~\cite{giebenhain2025pixel3dmm}, and FLAME~\cite{FLAME:SiggraphAsia2017} for topologically consistent facial geometry reconstruction.
We denote it \emph{naive FLAME}.
Specifically, after we distill the topology-consistent point map, \emph{i.e.} $\{X_i[q_{ij}^x][q_{ij}^y]\}_{j=1}^N$, from VGGT's raw prediction, we fit the FLAME parameters to minimize the error \emph{w.r.t} the predicted point cloud in each view:
\begin{equation}
    \mathcal{L} = \sum_{i=1}^V\sum_{j=1}^N v_{ij}\cdot||x_j-X_i[q_{ij}^x][q_{ij}^y]||_2^2
\end{equation}
Here, $\{x_j\}_{j=1}^N$ is the FLAME vertices generated by the given shape, expression, and pose parameters.
We minimize $\mathcal{L}$ by optimizing the FLAME parameters using the Adam optimizer.
As shown in Table~\ref{tab:comparison}, this baseline obtains inferior results as it is confined to the FLAME space.

\begin{figure}[t]
    \centering
    \includegraphics[width=0.475\textwidth]{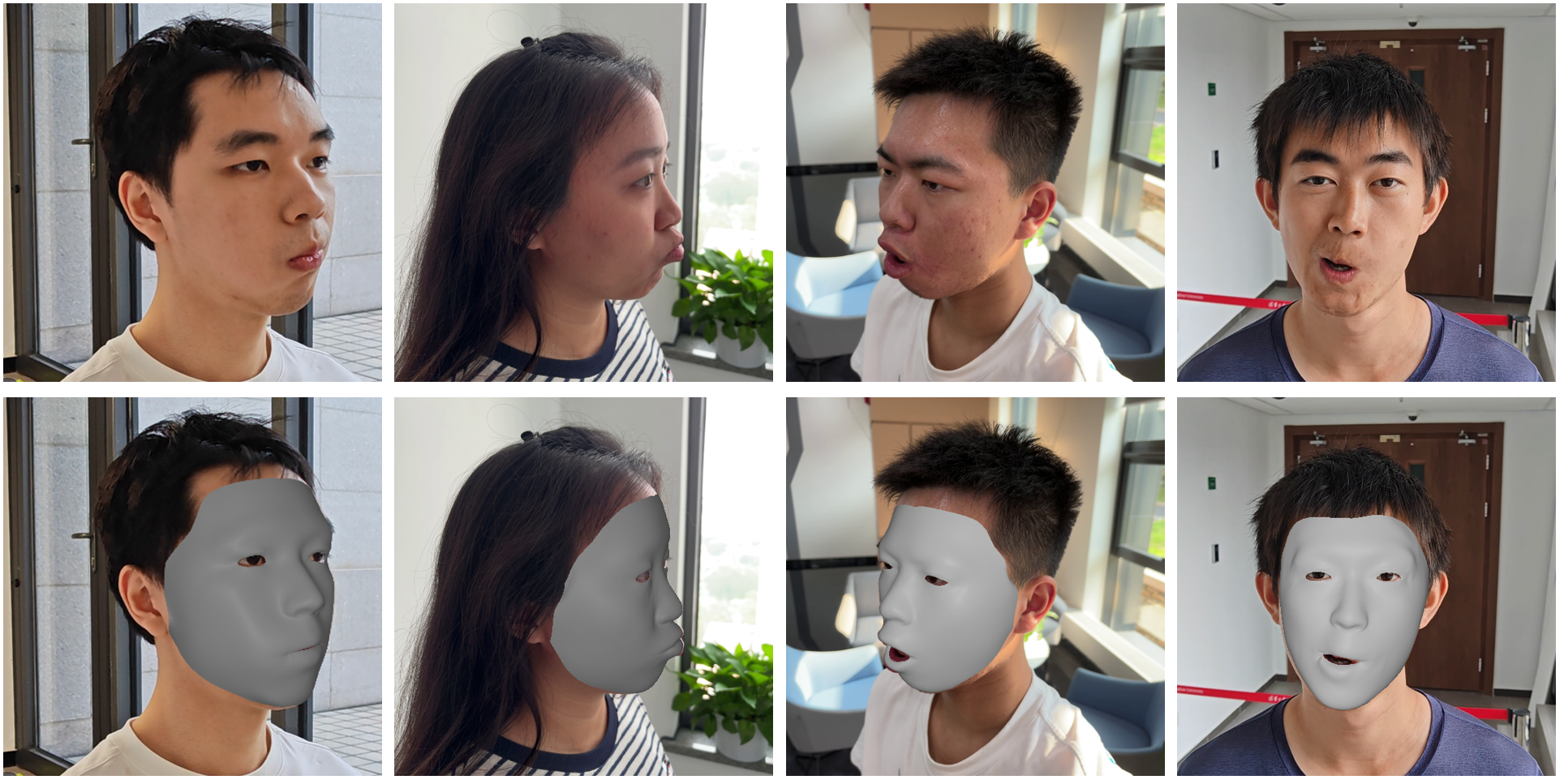}
    \caption{
        Overlay results of our method. We show one of the captured images in the first row and the corresponding geometry overlay results in the second row.
    }
    \label{Fig:overlay}
    \vspace{-0.5em}
\end{figure}

\subsection{Reconstruction Overlay}
After geometry reconstruction, we can re-render the mesh according to the reconstructed camera parameters of the captured images.
As shown in Figure~\ref{Fig:overlay}, our reconstructed mesh can well align the captured images.

\section{More Discussions on Failure Cases}
\begin{figure}[t]
    \centering
    \includegraphics[width=0.475\textwidth]{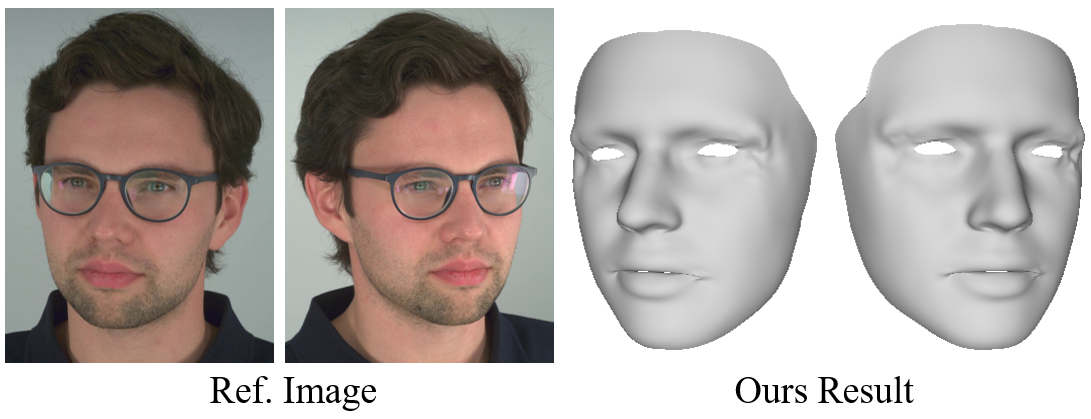}
    \caption{
        Failure case on subjects wearing glasses.
    }
    \label{Fig:glass}
\end{figure}

Although our method is robust to a wide range of in-the-wild capture conditions, it struggles when the subject wears glasses. As shown in Figure~\ref{Fig:glass}, eyeglasses violate the multi-view consistency assumption required by VGGT and TopBA, introducing strong specular reflections and semi-transparent regions around the eyes and causing noticeable artifacts in the fused geometry. In particular, we observe local surface depressions near the eyes and distorted eyelid shapes. Addressing eyeglass-related challenges would require explicitly modeling transparency and specular reflections, which is beyond the scope of our current approach.

\end{document}